\documentclass{article} 
\usepackage{iclr2027_conference,times}

\usepackage{amsmath,amsfonts,bm}

\def\eqref#1{equation~\ref{#1}}

\def\1{\bm{1}}

\DeclareMathAlphabet{\mathsfit}{\encodingdefault}{\sfdefault}{m}{sl}
\SetMathAlphabet{\mathsfit}{bold}{\encodingdefault}{\sfdefault}{bx}{n}

\usepackage{hyperref}
\usepackage{url}

\usepackage{graphicx}

\usepackage{amsmath}
\usepackage{amssymb}

\usepackage{booktabs}
\usepackage{multirow}
\usepackage{array}
\usepackage{tabularx}
\usepackage{adjustbox}

\usepackage{algorithm}
\usepackage{algorithmic}

\usepackage{xcolor}
\usepackage{tikz}
\usepackage{pgfplots}
\pgfplotsset{compat=1.18}

\usepackage{float}

\title{Lngram v2: Latent N-Gram Memory with Interpretable Discrete Representations}

\author{
Yunao Zheng\textsuperscript{1}
\quad
Bin Wen\textsuperscript{2,*,$\dagger$}
\quad
Xiaojie Wang\textsuperscript{1,$\dagger$}
\\[2pt]
Kaiyu Jiang
\quad
Xuanyu Zheng
\quad
Changyi Liu
\quad
Hongyi Fu
\\
Jianxiong Wang
\quad
Tianke Zhang
\quad
Haonan Fan
\quad
Yingxin Li
\\
Jiankang Chen
\quad
Xu Wang
\quad
Tingting Gao
\quad
Han Li
}

\iclrfinalcopy

\begin{document}

\maketitle

\begingroup
\renewcommand{\thefootnote}{}
\footnotetext{
\textsuperscript{1} Beijing University of Posts and Telecommunications.
\quad
\textsuperscript{2} Kuaishou Technology.
\quad
\textsuperscript{*} Project Leader.
\quad
\textsuperscript{$\dagger$} Corresponding authors: Xiaojie Wang and Bin Wen.
}
\endgroup

\begin{abstract}
Transformers lack a native lookup mechanism, requiring repeated dense computation to recognize and reuse local static patterns. Lngram v1 introduces tokenizer-independent conditional memory through discrete latent n-gram addressing, but its memory capacity is coupled with the backbone width, limiting scalability due to high parameter and activation costs. We propose Lngram v2, which decouples the number of routes, memory dimension, and backbone width, and introduces a context-aware grouped-query attention readout to scale memory capacity independently. A zero-value Sink and counterfactual surrogate gradients further improve readout selectivity and routing trainability while preserving hard discrete addressing. Experiments across vision--language models (VLMs) of different scales show consistent improvements, including successful scaling to a 30B-parameter model. Compared with Lngram v1, Lngram v2 substantially reduces both total and activated memory parameters while maintaining or improving language modeling performance. Further analysis shows that its discrete IDs preserve substantial semantic structure of continuous hidden states, enabling semantic recovery from IDs alone and stable ID--semantic associations across datasets. These results establish Lngram v2 as an efficient and scalable latent conditional memory mechanism whose discrete addresses also provide a structured interface for analyzing internal model representations.
\end{abstract}

\section{Introduction}

Transformers have become the dominant backbone for large language and multimodal models
\citep{vaswani2017attention,devlin2019bert,brown2020language,dosovitskiy2021image,radford2021learning,alayrac2022flamingo}.
However, attention and feed-forward networks must handle both context-dependent reasoning and operations closer to conditional lookup, such as recognizing common multi-token entities and recurring local patterns. Since standard Transformers lack a native lookup mechanism, such patterns must still be approximated through dense computation, consuming model capacity that could otherwise support compositional reasoning
\citep{cheng2026engram}.

To separate local pattern matching from the backbone, \citet{cheng2026engram} introduced conditional memory based on token-ID n-grams. Its addresses, however, depend on tokenizer IDs and fixed hashing, making the mechanism sensitive to tokenization boundaries, hash collisions, and difficult to extend uniformly beyond text. Lngram v1 instead discretizes hidden states into learnable symbols and constructs exact n-gram addresses in latent space, extending classical n-gram modeling
\citep{brants2007large}
to neural representations and enabling conditional memory over textual, visual, and cross-modal inputs.

Lngram v1 nevertheless has two limitations. First, its route count is coupled with backbone width, causing memory size and readout cost to increase with model dimension and limiting scalability. Second, although discrete Lngram IDs directly determine memory addresses, it remains unclear whether these IDs preserve the semantic structure of the continuous representations from which they are derived. Existing representation-analysis methods typically require continuous activations together with probes or learned dictionaries
\citep{alain2016understanding,hewitt2019structural,tenney2019bert,bau2017network,huben2024sparse}.
If Lngram IDs themselves retain semantic information, they may provide a useful interface for both memory addressing and representation analysis.

We therefore propose Lngram v2, which decouples route count, memory dimension, and backbone width so that memory capacity can scale independently. Retrieved memories of different n-gram orders are organized as memory tokens, and the current hidden state performs context-dependent readout through grouped-query attention (GQA)
\citep{ainslie2023gqa}.
A fixed zero-value Sink allows the model to suppress unreliable retrievals, while route-wise streaming controls intermediate activations at large scale. To train hard discrete routing, we use a counterfactual-lookup surrogate gradient that provides memory-dependent optimization signals while keeping the forward addressing path identical during training and inference, following the general principle of surrogate optimization for discrete decisions
\citep{bengio2013estimating}.

We further study the semantics encoded in Lngram's discrete representations using an ID-only semantic reader that receives only the multi-route discrete signature. The reader retains $65.77\%$--$84.27\%$ of the excess semantic readout capability of continuous hidden states. On held-out data, individual route codes exhibit reproducible associations with concepts such as people, animals, vehicles, and common objects, while most semantic capacity arises from distributed combinations across routes. These results suggest that Lngram IDs serve not only as memory addresses but also as a statistically analyzable discrete interface to internal representations.

We evaluate Lngram v2 on vision--language models at multiple scales. Under matched training data, compute budgets, and backbone optimization settings, it consistently improves Keye2B and Keye30B and scales successfully to a 30B-parameter model. Parameter-matched Sparse FFN and Product Key Memory (PKM) baselines show that the gains cannot be explained by additional parameters or sparse capacity alone. Compared with Lngram v1, Lngram v2 reduces the module's total parameters by $82.6\%$ and activated parameters per token by $95.2\%$, while maintaining or improving validation loss. Overall, Lngram v2 provides a scalable latent-space conditional memory mechanism together with a structured discrete interface for analyzing model representations.

\section{Architecture}

\subsection{Overview}

\begin{figure*}[t]
\centering
\includegraphics[width=0.8\linewidth]{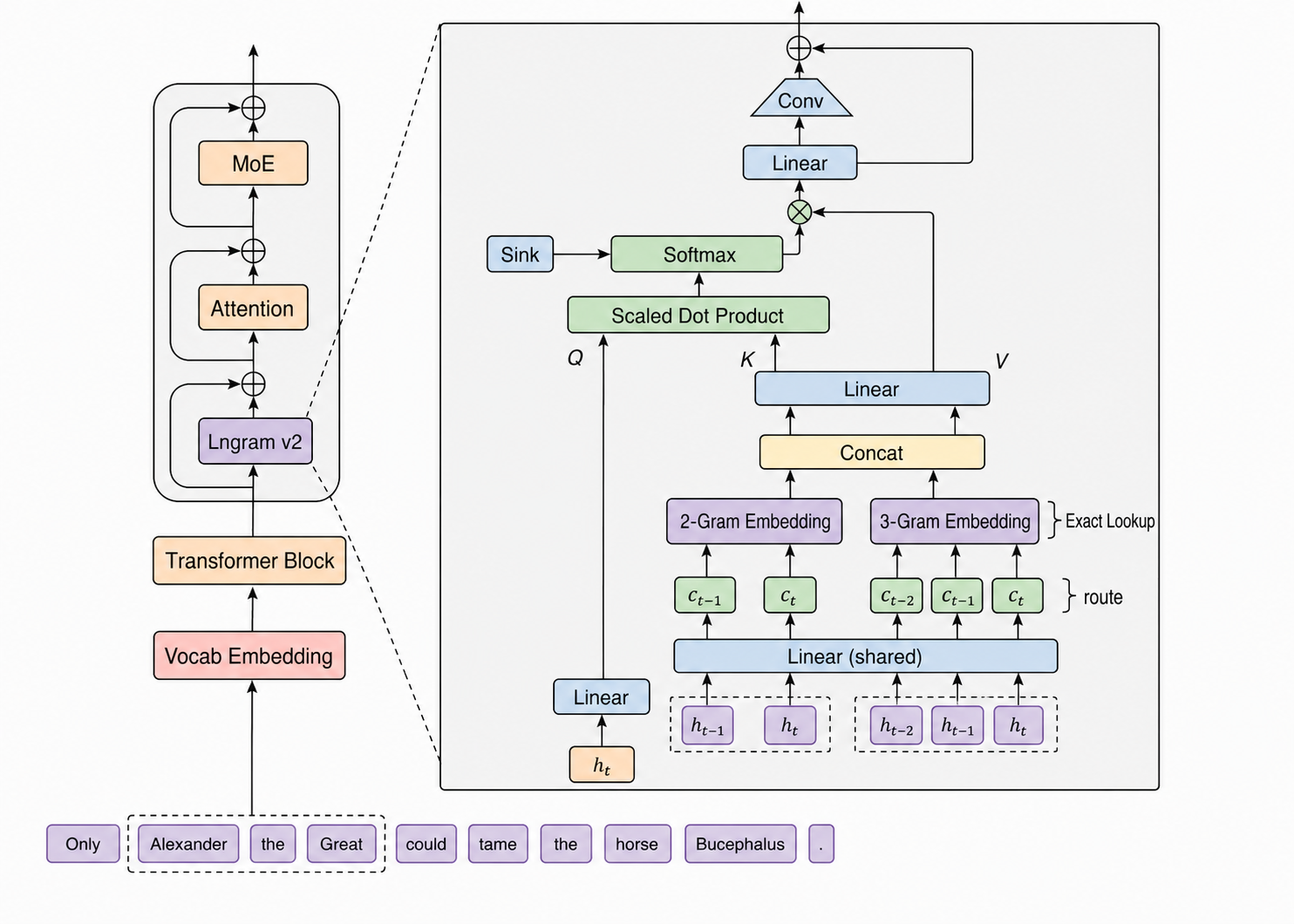}
\caption{Overall architecture of Lngram v2.}
\label{fig:lngram}
\end{figure*}

As illustrated in Figure~\ref{fig:lngram}, Lngram v2 is a conditional memory branch inserted into Transformer decoder layers to offload local pattern matching and storage from the backbone. Omitting the batch dimension, the input to layer $\ell$ is
\[
H^{(\ell)}=[h_1,\ldots,h_T]^\top\in\mathbb{R}^{T\times d}.
\]

Lngram v2 maps each hidden state to multiple discrete symbols, constructs local n-gram addresses for exact memory lookup, and organizes the retrieved entries as route-wise memory tokens. The current hidden state then performs context-dependent readout over these tokens through grouped-query attention (GQA), augmented with a fixed zero-value Sink that suppresses unreliable retrievals. The output is further processed by a short-range causal convolution and injected into the backbone as a residual.

The forward pass always uses hard discrete addressing, while the discretization projection is trained with counterfactual surrogate gradients. For large route counts, the readout can be streamed along the route dimension to reduce activation memory.

\subsection{Multi-Route Discrete Addressing}

Lngram v2 discretizes the hidden states through an independent projection:
\[
U=\mathrm{RMSNorm}(H),\qquad
Z=UW_q,\qquad
W_q\in\mathbb{R}^{d\times RM},
\]
where $R$ is the number of routes and $M$ is the number of bits per route. For position $t$, route $r\in\{0,\ldots,R-1\}$, and bit $j\in\{0,\ldots,M-1\}$,
\[
b_{t,r,j}=\mathbb{I}[z_{t,r,j}>0],
\]
and the bits are packed into a discrete symbol
\[
a_{t,r}
=
\sum_{j=0}^{M-1}b_{t,r,j}2^j,
\qquad
a_{t,r}\in\{0,\ldots,K-1\},
\quad K=2^M.
\]

Each hidden state is thus mapped to $R$ parallel discrete symbols. Since the addresses are derived from latent representations rather than tokenizer IDs, the same mechanism applies across modalities.

\subsection{Exact n-gram Retrieval}

For each order $n\in\mathcal{N}$, Lngram v2 maintains a memory table
\[
E^{(n)}\in\mathbb{R}^{RK^n\times d_m},
\]
where $d_m$ is the memory-entry dimension. For route $r\in\{0,\ldots,R-1\}$, the n-gram ending at position $t$ is encoded as
\[
g_{t,r}^{(n)}
=
rK^n+
\sum_{i=0}^{n-1}
a_{t-n+1+i,r}K^i,
\]
and retrieved exactly by
\[
m_{t,r}^{(n)}
=
E^{(n)}[g_{t,r}^{(n)}].
\]

Different routes occupy disjoint address ranges, yielding a unique entry for each route--n-gram combination without approximate search. Invalid prefixes and n-grams crossing packed-sequence boundaries are masked.

The retrieved entries of different orders are concatenated into
\[
x_{t,r}
=
\mathrm{Concat}_{n\in\mathcal{N}}
m_{t,r}^{(n)}
\in\mathbb{R}^{|\mathcal{N}|d_m},
\]
producing $R$ memory tokens per position.

\subsection{Context-Aware Readout}

Lngram v2 uses the current hidden state to contextually select among the retrieved route tokens. Let $H_q$ and $H_{kv}$ denote the numbers of query and KV heads, with head dimension $d_h$:
\[
H_qd_h=d,
\qquad
G=\frac{H_q}{H_{kv}}.
\]

The query and memory keys/values are
\[
Q_t
=
\mathrm{Reshape}
\left(
\mathrm{Norm}(h_t)W_Q
\right)
\in\mathbb{R}^{H_q\times d_h},
\]
\[
[K_{t,r};V_{t,r}]
=
\mathrm{Reshape}
\left(
\mathrm{Norm}(x_{t,r})W_{KV}
\right),
\qquad
K_{t,r},V_{t,r}\in\mathbb{R}^{H_{kv}\times d_h}.
\]

For query head $a\in\{0,\ldots,H_q-1\}$, define
\[
\kappa(a)=
\left\lfloor\frac{a}{G}\right\rfloor,
\]
and compute
\[
s_{t,a,r}
=
\frac{
Q_{t,a}^{\top}K_{t,r,\kappa(a)}
}{
\sqrt{d_h}
}.
\]

To avoid forcing probability mass onto unreliable memories, we introduce a zero-value Sink with fixed logit $b_0$:
\[
\alpha_{t,a,r}
=
\frac{
\exp(s_{t,a,r})
}{
\exp(b_0)+
\sum_{r'=0}^{R-1}\exp(s_{t,a,r'})
},
\qquad
o_{t,a}
=
\sum_{r=0}^{R-1}
\alpha_{t,a,r}V_{t,r,\kappa(a)}.
\]
We set
\[
b_0
=
\log
\frac{R(1-\mu_0)}{\mu_0},
\]
where $\mu_0$ specifies the initial total probability mass assigned to memory routes.

The head outputs are projected and refined by a depthwise causal convolution:
\[
\widetilde{v}_t
=
\mathrm{Concat}_{a=0}^{H_q-1}o_{t,a}\,W_O,
\]
and
\[
\widetilde{V}
=
[\widetilde{v}_1,\ldots,\widetilde{v}_T]^\top.
\]
The sequence representation is then refined as
\[
Y
=
\widetilde{V}
+
\mathrm{SiLU}
\left(
\mathrm{DWConv1D}
\left(
\mathrm{Norm}(\widetilde{V})
\right)
\right),
\]
with kernel size $4$. The result is injected as a pre-attention residual:
\[
H^{(\ell)}
\leftarrow
H^{(\ell)}+Y.
\]

For large $R$, the route dimension can be processed in blocks while accumulating the softmax normalization and value-weighted sum online, avoiding simultaneous storage of all route activations.

\subsection{Backpropagation Through Discrete Addressing}

The forward pass always uses hard decisions
\[
b_{t,r,j}
=
\mathbb{I}[z_{t,r,j}>0],
\]
which are non-differentiable. We therefore train the discretization projection using counterfactual memory lookups while preserving the hard forward path.

For one symbol in an n-gram window, let
\[
z=(z_0,\ldots,z_{M-1}),
\qquad
p_j=\sigma(\tau z_j).
\]
Holding the remaining symbols fixed, replacing this symbol by
$c\in\{0,\ldots,K-1\}$ yields a counterfactual retrieval $E^{(c)}$. Its probability is
\[
P(c\mid z)
=
\prod_{j=0}^{M-1}
p_j^{\beta_j(c)}
(1-p_j)^{1-\beta_j(c)},
\]
giving the conditional expectation
\[
\widetilde{E}(z)
=
\sum_{c=0}^{K-1}
P(c\mid z)E^{(c)}.
\]
This expectation is used only for the routing gradient; the actual forward output remains the hard lookup $E_{\mathrm{hard}}$.

In large-scale experiments, we use a one-bit approximation. For bit $j$, let $E_j^{(0)}$ and $E_j^{(1)}$ denote the counterfactual retrievals obtained by setting that bit to $0$ or $1$ while fixing all others. Given upstream gradient $g$,
\[
\frac{\partial\mathcal{L}}{\partial z_j}
\approx
\lambda\tau
p_j(1-p_j)
\left\langle
g,
E_j^{(1)}-E_j^{(0)}
\right\rangle.
\]

Only the backward gradient of the discretization projection is replaced: memory entries selected by hard lookup receive standard gradients, whereas counterfactual entries are used solely to estimate routing gradients. Thus, training and inference share exactly the same hard addressing and exact-lookup forward path. Further derivations and implementation details are provided in Appendix~\ref{app:surrogate-gradient}.

\section{Analysis}
\label{sec:analysis}

Lngram v2 maps continuous hidden states to discrete IDs for memory addressing. We investigate whether these IDs preserve the semantics of the original representations, whether such semantics can be recovered without accessing continuous hidden states, and whether they are localized to individual route codes or distributed across multiple routes.

\subsection{Experimental Setup}
\label{subsec:analysis_setup}

We analyze two Lngram layers of a 2B Keye2B+Lngram v2 model trained for approximately 20B additional tokens (Section~\ref{sec:experiments}). Each layer contains 128 routes with 16 codes per route, yielding a discrete signature
\[
C=(c_1,\ldots,c_{128})
\]
for each token.

We use 10,000 COCO 2017 images~\citep{lin2014microsoft}, split into 7,000 reference and 3,000 held-out images. All statistical criteria and readers are determined exclusively on the reference set and evaluated independently on the held-out set. We consider three semantic targets: caption mention, visual presence, and local presence, corresponding to concepts expressed in textual descriptions, the full image, and local image regions, respectively.

We first verify that the discrete space does not collapse. Across the two analyzed layers, each route uses approximately eight effective codes on average, the most frequent code accounts for about 30\% of occurrences, and all 4,096 theoretically possible route codes across the two layers are observed. Complete statistics are provided in Appendix~\ref{app:code_health}.

\subsection{Reading Hidden-State Semantics from Lngram IDs Alone}
\label{subsec:id_semantics}

Lngram v2 uses highly sparse discrete addressing while consistently improving downstream performance, suggesting that its discrete mapping may preserve stable semantic structure. While conventional representation analyses recover information from continuous hidden states using additional probes~\citep{alain2016understanding}, we ask whether the continuous representation can be removed entirely and its semantics recovered solely from Lngram IDs.

For each held-out query, we provide only its 128-route signature $C$ and retrieve the $K=10$ nearest distinct reference documents by route-Hamming distance. Nearest-neighbor prediction~\citep{cover1967nearest} is then used for caption mention, visual presence, and local presence. KNN based on the continuous hidden states $H_{\mathrm{model}}$ serves as the reference; we additionally report $Q_{\mathrm{route}}$ as a control and predictions using only nuisance information, such as image source and spatial position, as the baseline.

To quantify how much readable semantic information is retained after discretization, we define
\begin{equation}
\rho_{ID\leftarrow H}
=
\frac{
\overline{AP}_{ID}-\overline{AP}_{N}
}{
\overline{AP}_{H}-\overline{AP}_{N}
},
\end{equation}
where $N$ denotes the nuisance baseline.

\begin{table*}[t]
\centering
\caption{Semantic readout using only Lngram IDs.}
\label{tab:id_semantic}
\begin{tabular}{llccccc}
\toprule
Layer & Task &
$H_{\mathrm{model}}$ &
$Q_{\mathrm{route}}$ &
ID-KNN &
nuisance &
Excess AP Retained \\
\midrule
1  & caption mention & 0.239744 & 0.225916 & 0.200077 & 0.016789 & 82.21\% \\
1  & local presence  & 0.470371 & 0.450705 & 0.396941 & 0.008484 & 84.10\% \\
1  & visual presence & 0.317607 & 0.302142 & 0.273789 & 0.039000 & 84.27\% \\
13 & caption mention & 0.206947 & 0.177058 & 0.152670 & 0.016789 & 71.46\% \\
13 & local presence  & 0.403160 & 0.328081 & 0.268083 & 0.008493 & 65.77\% \\
13 & visual presence & 0.286478 & 0.250947 & 0.221730 & 0.039000 & 73.84\% \\
\bottomrule
\end{tabular}
\end{table*}

Table~\ref{tab:id_semantic} shows that discrete IDs recover a substantial fraction of the readable semantics in continuous hidden states. Layer~1 retains 82.21\%--84.27\% of the excess AP of $H_{\mathrm{model}}$, while Layer~13 retains 65.77\%--73.84\%. All six settings exhibit the consistent ordering
\[
H_{\mathrm{model}}>Q_{\mathrm{route}}>\mathrm{ID}\gg N,
\]
indicating that discretization incurs some information loss but preserves most of the readable semantic structure.

To exclude nuisance-driven correlations, we additionally perform 500 document-level permutations of the reference signatures under the same nuisance conditions. In all six settings, the observed AP exceeds every permuted result. Using the Monte Carlo $p$-value for random permutation tests~\citep{phipson2010permutation}, all settings yield
\[
p=1/501=0.001996.
\]
Thus, the ID-only readout cannot be explained by biases such as image source or spatial position.

We next ask whether semantic information can already be localized to individual route codes. For a code
\[
u=(l,r,c),
\]
we estimate its association with semantic concepts after controlling for nuisance variables. Candidate associations and their effect directions are determined strictly on the reference set and independently evaluated on the held-out set. Among 16,304 candidate associations discovered on the reference set, 14,002 are evaluable on the held-out set, of which 9,220 satisfy consistent effect direction, confidence-interval, and Benjamini--Hochberg multiple-testing criteria~\citep{benjamini1995controlling}. This corresponds to an overall replication rate of 65.85\%; for local presence, the rates reach 83.01\% and 84.94\% in Layers~1 and~13, respectively.

These results show that individual route codes can form reproducible concept-level associations. Appendix~\ref{app:id_dictionary} provides 15 representative examples. For instance,
\[
C_{1,75}=10
\]
is associated with \emph{giraffe}: on the held-out set, the conditional probability increases from a nuisance-controlled baseline of 2.37\% to 21.70\%, a gain of 19.33 percentage points. Similarly,
\[
C_{13,53}=14
\]
is associated with \emph{person}, whose conditional probability increases from 53.60\% to 63.25\%. These examples illustrate the discovered structure, while all reported replication and significance statistics follow the complete reference--held-out validation protocol.

Overall, Lngram IDs already form a queryable semantic interface: the complete signature recovers much of the semantics encoded in the corresponding hidden state without accessing its continuous representation, while individual route codes exhibit stable and interpretable concept associations.

\subsection{Distributed Semantic Representation Across Route Codes}
\label{subsec:distributed_semantics}

Although individual route codes exhibit stable semantic preferences, the complete signature provides substantially stronger readout capability. To determine whether this gap arises from joint encoding across routes, we recombine the marginal semantic effects of individual codes estimated on the reference set and measure how much semantic information can be recovered using different numbers of routes. This analysis uses only discrete IDs on the held-out set, without continuous representations or additional fitted parameters; details are provided in Appendix~\ref{app:distributed_semantics}.

Simply accumulating the marginal semantic effects of individual codes retains 31.79\%--54.82\% of the excess AP of $H_{\mathrm{model}}$ across the six settings, confirming that individual codes contain composable semantic information. However, this remains substantially below the 65.77\%--84.27\% achieved by ID-KNN using the complete signature. Moreover, even when routes are prioritized by their contribution to each query--concept pair, using all 128 routes consistently outperforms using only the single best route, with AP ratios ranging from $2.02$ to $12.32$.

These results indicate that Lngram semantics are not concentrated in a small number of specialized routes, but are distributed across joint route configurations, consistent with distributed representations in neural networks~\citep{hinton1986distributed}. Individual route codes provide local and interpretable semantic tendencies, while the complete signature captures richer compositional structure across routes. Lngram IDs therefore serve not only as memory addresses but also as a statistically analyzable, queryable, and reproducible semantic interface to internal model representations.

\section{Experiments}
\label{sec:experiments}

This section evaluates the effectiveness, architectural design, and parameter efficiency of Lngram v2. We first evaluate Lngram v2 on the 2B and 30B variants of Keye2-VL~\citep{keyevl2}, with additional comparisons against Product Key Memory (PKM) and a parameter-matched Sparse FFN on the 2B model. We then study the main architectural hyperparameters, the Sink, and inference efficiency. Finally, we compare Lngram v2 with v1 to assess whether v2 improves parameter efficiency while preserving or improving language modeling performance.

\subsection{Main Results}
\label{sec:vlm_experiments}

We evaluate Lngram v2 on Keye2B and Keye30B. All models continue training from the same respective base models for approximately 5B tokens with identical training data and backbone optimization settings. Lngram v2 is inserted into a shallow and an intermediate layer and uses 4-bit discrete codes with 2/3-gram memory. On Keye2B, it adds approximately 155.6M parameters and 20.9M activated parameters per token, corresponding to $6.11\%$ and $0.82\%$ of the base model, respectively. On Keye30B, the corresponding increases are approximately $2.0\%$ and $0.85\%$. Detailed configurations are provided in Appendix~\ref{sec:hyperparameters}. For the Keye2B Baseline and Lngram v2 results, we use three seeds (19260818, 19260819, and 19260820) and report the mean $\pm$ standard deviation.

We evaluate on public vision--language benchmarks including OCRBench~\citep{liu2024ocrbench}, MathVista~\citep{lu2024mathvista}, MM-Vet~\citep{yu2024mmvet}, HallusionBench~\citep{guan2024hallusionbench}, MMStar~\citep{chen2024mmstar}, MMMU~\citep{yue2024mmmu}, AI2D~\citep{kembhavi2016diagram}, and MMBench~\citep{liu2024mmbench}.

For architecture-level comparison with existing memory mechanisms, we additionally evaluate Product Key Memory (PKM)~\citep{lample2019large} on Keye2B. Conditional memory methods such as Engram rely on text token IDs and are tied to a particular vocabulary, making them difficult to apply directly to visual inputs~\citep{cheng2026engram}, whereas Lngram is designed to support arbitrary modalities. We also construct a parameter-matched Sparse FFN to determine whether the gains can be explained by additional parameters or sparse computational capacity. Specifically, a top-1 Sparse SwiGLU residual branch~\citep{shazeer2020glu,fedus2022switch,dai2024deepseekmoe} is inserted into the same 2nd and 14th decoder layers as Lngram v2, without discrete addressing, n-grams, memory tables, or lookup. It adds 155.845M parameters and 20.972M activated parameters per token, closely matching Lngram v2.

\begin{table*}[t]
\centering
\caption{Comparison results on Keye2B and Keye30B. Keye2B Baseline and Lngram v2 results are reported as mean $\pm$ standard deviation over three seeds; other entries are single-run results.}
\label{tab:vlm_lngram_v2}
\resizebox{\textwidth}{!}{
\begin{tabular}{llccccccccccc}
\toprule
Backbone & Model & OCR & MathVista & MMVet & Hallusion & MMStar & MMMU & AI2D & MMB-EN & MMB-CN & Bench V2.1 & Avg \\
\midrule
\multirow{4}{*}{Keye2B}
& Baseline
& $65.07 \pm 0.26$
& $50.23 \pm 0.82$
& $37.35 \pm 0.65$
& $60.43 \pm 0.36$
& $45.09 \pm 0.25$
& $32.48 \pm 1.01$
& $58.57 \pm 0.32$
& $52.97 \pm 0.76$
& $33.77 \pm 1.17$
& $36.51 \pm 0.69$
& $47.25 \pm 0.15$ \\

& + PKM
& 65.40 & 49.10 & \textbf{39.22} & 60.04 & 45.20 & 29.78
& 57.87 & \textbf{53.33} & \textbf{36.15} & 36.31 & 47.24 \\

& + Matched Sparse FFN
& 64.30 & \textbf{51.90} & 32.48 & 60.46 & 44.13 & \textbf{33.11}
& 58.00 & 48.14 & 29.80 & 36.31 & 45.86 \\

& + Lngram v2
& $\mathbf{66.47} \pm 0.21$
& $50.57 \pm 1.05$
& $39.01 \pm 0.44$
& $\mathbf{60.99} \pm 0.31$
& $\mathbf{45.64} \pm 0.08$
& $32.37 \pm 0.55$
& $\mathbf{59.39} \pm 0.69$
& $53.15 \pm 0.45$
& $34.86 \pm 1.22$
& $\mathbf{36.85} \pm 0.75$
& $\mathbf{47.93} \pm 0.18$ \\
\midrule
\multirow{2}{*}{Keye30B}
& Baseline
& \textbf{81.30} & \textbf{80.90} & 69.45 & 74.03 & 69.60 & 67.78
& 82.09 & 83.75 & 81.73 & 91.66 & 78.23 \\

& + Lngram v2
& 81.00 & 80.80 & \textbf{72.02} & \textbf{74.87} & \textbf{70.93}
& \textbf{70.22} & \textbf{83.97} & \textbf{86.92}
& \textbf{83.75} & \textbf{92.62} & \textbf{79.71} \\
\bottomrule
\end{tabular}
}
\end{table*}

As shown in Table~\ref{tab:vlm_lngram_v2}, Lngram v2 improves the average score from $47.25 \pm 0.15$ to $47.93 \pm 0.18$ on Keye2B, corresponding to a mean gain of 0.68 points across three seeds. The improvement is consistent across all three seeds. PKM reaches 47.24, close to the Baseline mean, while Lngram v2 exceeds PKM by 0.69 points. In contrast, the parameter-matched Sparse FFN obtains only 45.86 despite nearly identical numbers of additional total and activated parameters. These results indicate that the gains of Lngram v2 cannot be simply attributed to additional parameters or sparse computational capacity.

On Keye30B, Lngram v2 improves the ten-benchmark average from 78.23 to 79.71, a gain of 1.48 points, demonstrating that the method scales to the 30B regime. Given the substantially higher training cost at this scale and the controlled Keye2B comparisons above, we do not repeat the PKM and Matched Sparse FFN baselines on Keye30B.

Paired statistical tests on Keye30B yield a 95\% confidence interval of $[0.86,2.11]$ for the average improvement. After Holm~\citep{holm1979simple} correction, AI2D and MMBench-EN remain statistically significant, while MMBench-CN is significant under Benjamini--Hochberg (BH)~\citep{benjamini1995controlling} correction. Complete results are provided in Appendix~\ref{app:vlm_significance}.

Lngram v2 also maintains lower LM loss than the Baseline during the middle and later stages of training across different route configurations, with routes=64 achieving the lowest loss. This suggests that the discrete memory branch can participate stably in backbone optimization. Training curves are provided in Figure~\ref{fig:counterfactual_vs_ste}.

\subsection{Ablation Studies}
\label{sec:ablation_v2}

We investigate the main architectural hyperparameters of Lngram v2, where $r$ denotes the number of routes, $m$ the bits per route, KV the number of KV heads, and mem the memory-entry dimension. Because parameter counts vary across configurations, we focus on settings with similar total parameter counts or small architectural differences. Evaluation is performed on OCRBench~\citep{liu2024ocrbench}, CountBenchQA~\citep{paiss2023countclip,beyer2024paligemma}, and RealWorldQA~\citep{xai2024grok15v}.

\begin{table*}[t]
\centering
\caption{Hyperparameter ablations for Lngram v2.}
\label{tab:lngram_v2_ablation}
\scriptsize
\setlength{\tabcolsep}{3.5pt}
\resizebox{\textwidth}{!}{
\begin{tabular}{lcccccc}
\toprule
Configuration
& Total Params
& Active Params/token
& OCRBench $\uparrow$
& CountBenchQA $\uparrow$
& RealWorldQA $\uparrow$
& Avg $\uparrow$ \\
\midrule
Baseline
& -- & --
& 65.30 & 59.55 & 36.08 & 53.64 \\
r16/m4/KV8
& 35.939M & 18.121M
& 65.90 & 57.49 & 39.22 & 54.20 \\
r64/m4/KV8
& 90.203M & 18.932M
& 64.90 & 59.75 & \textbf{42.48} & 55.71 \\
r23/m4/KV8/mem256
& 89.400M & 38.175M
& 66.00 & 60.16 & 37.91 & 54.69 \\
r512/m4/KV8/mem256
& 1187.014M & 46.687M
& 64.70 & 59.55 & 41.18 & 55.14 \\
r16/m5/KV2
& 155.804M & 17.400M
& 66.20 & 57.29 & 35.42 & 52.97 \\
r123/m4/KV2
& 156.115M & 19.143M
& 66.10 & 56.26 & 38.69 & 53.69 \\
r121/m4/KV16
& 155.689M & 20.944M
& \textbf{66.70} & \textbf{60.99} & 40.78 & \textbf{56.16} \\
\bottomrule
\end{tabular}
}
\end{table*}

Table~\ref{tab:lngram_v2_ablation} shows that increasing the number of routes is generally effective with modest activation overhead. With $m=4$ and KV8 fixed, increasing routes from r16 to r64 improves the average from 54.20 to 55.71, while activated parameters increase only from 18.121M to 18.932M. Moreover, r64/m4/KV8 and r23/m4/KV8/mem256 both contain about 90M total parameters, but the former performs 1.02 points better with roughly half the activated parameters. Thus, under a similar parameter budget, increasing the number of discrete routes is more effective than increasing the memory dimension alone.

Readout capacity is also important: r121/m4/KV16 achieves the best average score of 56.16. By contrast, r512/m4/KV8/mem256 adds more than 1B parameters but remains below r64/m4/KV8, showing diminishing returns from simply enlarging memory capacity. Most configurations outperform the Baseline, indicating that the gains are robust across architectural settings.

We further ablate the zero-value Sink on Keye30B. The ten-benchmark average increases from 79.39 without the Sink to 79.71 with it; MMBench-EN improves from 85.91 to 86.92 and Benchmark V2.1 from 91.66 to 92.62. We therefore enable the Sink by default.

\subsection{Efficiency Evaluation}
\label{sec:inference_efficiency_v2}

We measure prefill latency at sequence lengths of 256, 512, and 1024 and single-token decode latency on one H800 GPU. Complete results are provided in Appendix~\ref{app:efficiency_detail}.

Moderate-scale Lngram v2 configurations incur only modest inference overhead. For r64/m4/KV8, which improves the three-benchmark average by 2.07 points over the Baseline, prefill latency at length 1024 increases from 56.99 ms to 60.49 ms ($+6.1\%$), with memory usage increasing from 5.24 GiB to 5.46 GiB. Decode latency increases from 48.26 to 52.14 ms/token ($+8.0\%$), with only 0.16 GiB additional memory. The lighter r16/m4/KV8 configuration incurs approximately $3.7\%$ prefill overhead at length 1024.

These results show that Lngram v2 improves performance without substantial inference-time computation, while its route count and readout capacity provide a flexible trade-off between model capability and deployment cost.

\subsection{Comparison with Lngram v1}
\label{sec:v1_v2_comparison}

Lngram v2 is also designed to improve the parameter efficiency of conditional memory. In Lngram v1, the route count is coupled with the backbone representation dimension, resulting in large memory tables and readout modules. Lngram v2 decouples the route count from backbone width, allowing memory capacity to scale independently.

We compare the two designs using the same 0.6B DeepSeek-V2-style sparse mixture-of-experts (MoE) model~\citep{deepseekai2024deepseekv2,dai2024deepseekmoe}. All models are trained from random initialization for approximately 1.31B tokens on FineWeb-Edu~\citep{penedo2024fineweb}, using the same data order, optimizer, and random seed. Detailed configurations are provided in Appendix~\ref{app:v1v2_setup}.

\begin{table}[t]
\centering
\caption{Validation losses of Baseline, Lngram v1, and Lngram v2.}
\label{tab:v1_v2_comparison}
\begin{tabular}{lccccc}
\toprule
Model & Baseline & Lngram v1 & Lngram v2 & Lngram v2 & Lngram v2 \\
\midrule
Routes & -- & -- & 192 & 64 & 16 \\
Val Loss $\downarrow$ & 2.8899 & 2.8609 & \textbf{2.8527} & 2.8578 & 2.8584 \\
\bottomrule
\end{tabular}
\end{table}

As shown in Table~\ref{tab:v1_v2_comparison}, both Lngram v1 and all v2 configurations outperform the memory-free Baseline. At $R=192$, where the total parameter count is approximately matched to v1, Lngram v2 further reduces validation loss from 2.8609 to 2.8527, showing that the improved parameter efficiency does not sacrifice modeling capability.

More importantly, at $R=16$, Lngram v2 reduces the total parameter count of the Lngram module by $82.6\%$ and the activated parameters per token by $95.2\%$ relative to v1, while achieving a slightly lower validation loss (2.8584 vs.\ 2.8609). Lngram v2 therefore preserves the benefits of conditional memory at substantially lower parameter and activation costs, making it more suitable for scaling to larger models.

\section{Conclusion}

We introduce Lngram v2, an efficient and scalable conditional memory mechanism in the latent space. By decoupling the number of routes, memory dimension, and backbone width and employing a context-aware readout, Lngram v2 substantially reduces the parameter and activation overhead of conditional memory, enabling it to scale to larger models while maintaining consistent performance gains. Further mechanistic analysis shows that Lngram's discrete IDs not only serve as memory addresses but also systematically preserve the representational structure and semantic information of continuous hidden states, forming stable and reproducible ID--semantic associations. Lngram v2 therefore provides, on the one hand, low-cost and scalable discrete memory for large models and, on the other hand, transforms continuous internal representations into a statistically analyzable and queryable discrete interface, offering a new avenue for analyzing and understanding the internal representations of large models.

\bibliographystyle{iclr2027_conference}
\bibliography{refs}

\newpage
\appendix

\section{Detailed Derivation of the Counterfactual Surrogate Gradient}
\label{app:surrogate-gradient}

\subsection{Notation and Problem Setup}

Consider a fixed n-gram order $n$, a fixed route $r$, and a local position within the n-gram ending at position $t$,
\[
u\in\{0,\ldots,n-1\},
\]
where $u=0$ denotes the earliest position in the window. Let the bit logits at this position be
\[
z=(z_0,\ldots,z_{M-1})\in\mathbb{R}^M,
\qquad
p_j=\sigma(\tau z_j),
\]
where $\tau>0$ is a temperature coefficient.

When constructing the surrogate, we fix all local symbols in the n-gram except the current one, as well as the symbols on all other routes. For any local symbol
\[
c\in\{0,\ldots,K-1\},
\qquad
K=2^M,
\]
let $\beta(c)\in\{0,1\}^M$ denote its binary representation, with $\beta_j(c)$ denoting its $j$-th bit.

After replacing the symbol at the current position with $c$, the corresponding counterfactual address is
\[
g_{t,r}^{(n)}(u\leftarrow c)
=
rK^n
+
\sum_{\substack{i=0\\i\neq u}}^{n-1}
a_{t-n+1+i,r}K^i
+
cK^u.
\]

The corresponding counterfactual retrieval vector is
\[
E_c
=
E^{(n)}
\left[
g_{t,r}^{(n)}(u\leftarrow c)
\right]
\in\mathbb{R}^{d_m}.
\]

Let the upstream gradient received by the retrieved vector be
\[
g
=
\frac{\partial\mathcal{L}}{\partial E}
\in\mathbb{R}^{d_m}.
\]
Here, $g$ already incorporates the gradients propagated through the subsequent GQA, zero-value Sink, output projection, and convolutional branch. The surrogate therefore only needs to handle the local discrete dependency from the retrieved result to the routing logits.

\subsection{Local Exact Surrogate}

Here, ``exact'' refers to the analytical gradient of the local conditional-expectation surrogate, rather than the gradient of the original hard-lookup path. We define the local conditional expected retrieval vector as
\[
\mu(z)
=
\sum_{c=0}^{K-1}
P(c\mid z)E_c,
\]
where
\[
P(c\mid z)
=
\prod_{j=0}^{M-1}
p_j^{\beta_j(c)}
(1-p_j)^{1-\beta_j(c)}.
\]

This is the probability that the local symbol takes value $c$ when the $M$ bits at the current position are treated as independent Bernoulli variables. Importantly, this expectation is applied only to one local position in the n-gram window. The other $n-1$ symbols always retain their hard discrete values from the actual forward pass. Consequently, only $K$ candidate symbols need to be enumerated, rather than all $K^n$ possible n-grams.

We define the local surrogate objective as
\[
\mathcal{L}_{\mathrm{local}}(z)
=
\langle g,\mu(z)\rangle
=
\sum_{c=0}^{K-1}
P(c\mid z)\langle g,E_c\rangle.
\]

To derive its gradient with respect to $z_j$, we first write
\[
\log P(c\mid z)
=
\sum_{j=0}^{M-1}
\left(
\beta_j(c)\log p_j
+
(1-\beta_j(c))\log(1-p_j)
\right).
\]

Since
\[
p_j=\sigma(\tau z_j),
\]
we have
\[
\frac{\partial p_j}{\partial z_j}
=
\tau p_j(1-p_j).
\]

It follows that
\[
\frac{\partial\log P(c\mid z)}{\partial z_j}
=
\tau\bigl(\beta_j(c)-p_j\bigr),
\]
and therefore
\[
\begin{aligned}
\frac{\partial P(c\mid z)}{\partial z_j}
&=
P(c\mid z)
\frac{\partial\log P(c\mid z)}{\partial z_j}
\\
&=
\tau
P(c\mid z)
\bigl(\beta_j(c)-p_j\bigr).
\end{aligned}
\]

Substituting this into the definition of $\mathcal{L}_{\mathrm{local}}$ gives
\[
\begin{aligned}
\frac{\partial\mathcal{L}_{\mathrm{local}}}{\partial z_j}
&=
\sum_{c=0}^{K-1}
\frac{\partial P(c\mid z)}{\partial z_j}
\langle g,E_c\rangle
\\
&=
\tau
\sum_{c=0}^{K-1}
P(c\mid z)
\bigl(\beta_j(c)-p_j\bigr)
\langle g,E_c\rangle.
\end{aligned}
\]

In practice, we further multiply this expression by a global surrogate scaling coefficient $\lambda$:
\[
\boxed{
\frac{\partial\mathcal{L}}{\partial z_j}
\approx
\lambda\tau
\sum_{c=0}^{K-1}
P(c\mid z)
\bigl(\beta_j(c)-p_j\bigr)
\langle g,E_c\rangle
}.
\]

This is the analytical form of the local exact surrogate. In implementation, the above gradient is computed for every local position $u$ in every valid n-gram window and accumulated into the routing logits of the corresponding positions. Because the same routing logit may participate in n-grams ending at multiple positions and in n-grams of multiple orders, its final surrogate gradient is the sum of the contributions from all relevant windows.

When $M=4$,
\[
K=2^M=16,
\]
only $16$ counterfactual symbols need to be enumerated for each local position, making the computation practically manageable.

\subsection{One-Bit Approximate Surrogate}

When computational efficiency is more important, we further restrict the counterfactual to a single bit at a time. Let the hard symbol produced by the actual forward pass at the current position be
\[
\hat{c}
=
\sum_{j=0}^{M-1}\hat{b}_j2^j,
\]
where
\[
\hat{b}_j=\mathbb{I}[z_j>0].
\]

For bit $j$, we keep all other bits of the symbol and all other symbols in the window fixed, and force only this bit to $0$ or $1$. Denote the resulting two counterfactual symbols by
\[
\hat{c}_j^{(0)},
\qquad
\hat{c}_j^{(1)},
\]
with corresponding retrievals
\[
E_j^{(0)},
\qquad
E_j^{(1)}.
\]

We define the local counterfactual score for this bit as
\[
s_j
=
\left\langle
g,
E_j^{(1)}-E_j^{(0)}
\right\rangle.
\]

This quantity measures how much the retrieved result changes along the current upstream-gradient direction when only bit $j$ is switched from $0$ to $1$, while all other discrete decisions remain fixed.

In implementation, $E_j^{(0)}$ and $E_j^{(1)}$ need not be constructed explicitly as two separate retrievals. Let $E_{\mathrm{hard}}$ denote the retrieval from the actual forward pass and $E_{\mathrm{flip}}$ the retrieval obtained after flipping the $j$-th bit of the current hard address. If the current bit is $0$, then
\[
E_{\mathrm{hard}}=E_j^{(0)},
\qquad
E_{\mathrm{flip}}=E_j^{(1)};
\]
if the current bit is $1$, then
\[
E_{\mathrm{hard}}=E_j^{(1)},
\qquad
E_{\mathrm{flip}}=E_j^{(0)}.
\]

Thus, the score can be written uniformly as
\[
s_j
=
(1-2\hat{b}_j)
\left\langle
g,
E_{\mathrm{flip}}-E_{\mathrm{hard}}
\right\rangle
=
\left\langle
g,
E_j^{(1)}-E_j^{(0)}
\right\rangle.
\]

Using the local slope of the sigmoid to characterize the sensitivity of the bit to the continuous logit gives the one-bit approximate surrogate:
\[
\boxed{
\frac{\partial\mathcal{L}}{\partial z_j}
\approx
\lambda\tau
p_j(1-p_j)
\left\langle
g,
E_j^{(1)}-E_j^{(0)}
\right\rangle
}.
\]

Compared with the exact surrogate, this approximation no longer enumerates all $K$ symbols and instead requires only one flipped counterfactual address per bit. The counterfactual lookup complexity per local position is therefore reduced from
\[
O(K)
\]
to
\[
O(M).
\]

When $M=4$, the number of counterfactual candidates per local position is reduced from $16$ to $4$, substantially reducing the additional lookup overhead during training.

\subsection{Correspondence to the Actual Implementation}

The derivation above considers a single n-gram, a single route, and a single local position within the window. The actual implementation performs the same computation in blocks over n-gram orders and routes.

In full-lookup mode, for each order $n$, the hard lookup first produces a tensor of shape
\[
[B,T,R,d_m].
\]
The surrogate operation is applied directly to the retrieval tensor of that order. Its numerical value is kept exactly unchanged in the forward pass, while the corresponding upstream gradient
\[
g\in\mathbb{R}^{d_m}
\]
is accessed during backpropagation and used to query the same n-gram table at counterfactual addresses. After surrogate processing is completed for all orders, the resulting tensors are concatenated along the last dimension, yielding the route memory tensor passed to the subsequent GQA:
\[
[B,T,R,|\mathcal{N}|d_m].
\]

In streaming mode, the computational logic remains unchanged, except that the route dimension is divided into smaller chunks. For the current interval
\[
[\text{route}_{\mathrm{start}},
\text{route}_{\mathrm{end}}),
\]
exact n-gram lookup and the surrogate are first performed for the corresponding routes. The retrieved results are then projected into keys and values, while the GQA normalization term and value-weighted sum are accumulated using an online softmax. The surrogate formula for each route chunk is identical to that in full-lookup mode, and its gradients are automatically accumulated into the complete routing logits.

Under context-parallel training, different positions may reside on different devices. The implementation therefore first gathers the limited amount of historical route information required to construct local n-grams and then explicitly forms local logit windows of shape
\[
[B,T,n,R,M].
\]
For each local target position, all other symbols in the window remain fixed, while only one local symbol is replaced or one bit is flipped, using exactly the same exact or approximate surrogate. Context parallelism changes only how the historical window is obtained; it does not alter the definition of the counterfactual gradient.

Importantly, the surrogate replaces only the backward gradient of the routing logits and does not modify the actual forward computation. For the hard lookup result $E_{\mathrm{hard}}$, its upstream gradient is still propagated according to the standard chain rule, so the memory-table entry selected by the actual lookup receives its normal parameter gradient. In contrast, the table parameters accessed by counterfactual queries are treated as constants during surrogate computation. Therefore, $E_c$, $E_j^{(0)}$, and $E_j^{(1)}$ do not propagate additional gradients into the corresponding counterfactual table entries.

The training procedure of Lngram v2 can therefore be summarized as
\[
\underbrace{
\text{hard routing}
\rightarrow
\text{exact address}
\rightarrow
\text{hard table lookup}
}_{\text{actual forward path}},
\]
while only the backward pass uses
\[
\underbrace{
\text{counterfactual lookup}
\rightarrow
\text{surrogate gradient}
}_{\text{surrogate gradient for routing logits}}.
\]

This design guarantees exactly the same hard discrete addressing behavior during training and inference, while the counterfactual computation is used solely to provide structure-aware optimization signals for the non-differentiable routing decisions. At inference time, the surrogate gradient is unnecessary; only hard discretization, exact n-gram lookup, and the subsequent readout computation are retained.

\subsection{Direct STE Fails to Effectively Optimize Discrete Routing}
\label{app:a7}

The discrete routing and exact address lookup of Lngram v2 are non-differentiable, making gradient estimation for the discrete choices critical to optimization. To verify the necessity of the surrogate gradient in Lngram v2, we conduct a strictly controlled ablation in which the model architecture, forward computation, training data, and optimization configuration are kept identical, and only the original surrogate gradient is replaced with a direct Straight-Through Estimator (STE). Both methods perform the same hard quantization and exact table lookup in the forward pass. The difference lies in the backward pass: direct STE passes gradients through the discrete address and approximates the binary threshold gradient using the local slope of the sigmoid, without accounting for the actual memory contents retrieved after switching to a different discrete address. In contrast, the surrogate gradient of Lngram v2 constructs its gradient from the actual lookup results corresponding to candidate discrete choices.

\begin{figure}[t]
    \centering
    \includegraphics[width=0.90\linewidth]
    {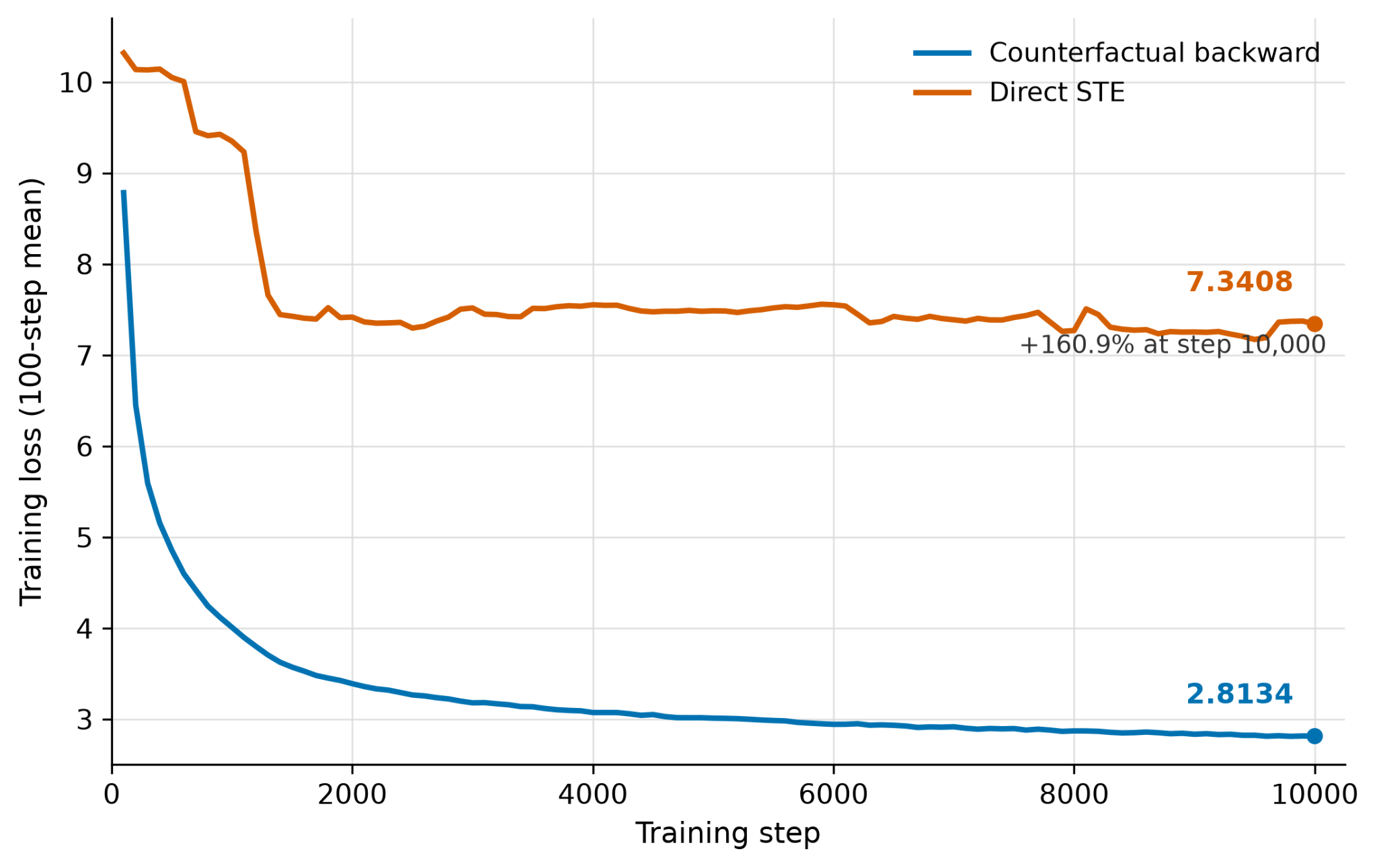}
    \caption{Comparison of training loss.}
    \label{fig:counterfactual_vs_ste}
\end{figure}

The experiment uses the $R=16$ configuration of Lngram v2, with Lngram inserted into the 2nd and 12th Transformer layers. Each route uses 4-bit discrete codes and 2/3-gram memory, with a memory dimension of 64. Both experiments are trained in BF16 on 8 GPUs with a global batch size of 32, an initial learning rate of $3\times10^{-4}$, a warmup ratio of $1\%$, and a random seed of 42. We retain the original cosine learning-rate schedule over 61,036 steps but stop training early at step 10,000. Figure~\ref{fig:counterfactual_vs_ste} reports the average training loss over each consecutive block of 100 training steps without additional smoothing.

\begin{figure}[t]
    \centering
    \includegraphics[width=0.90\linewidth]
    {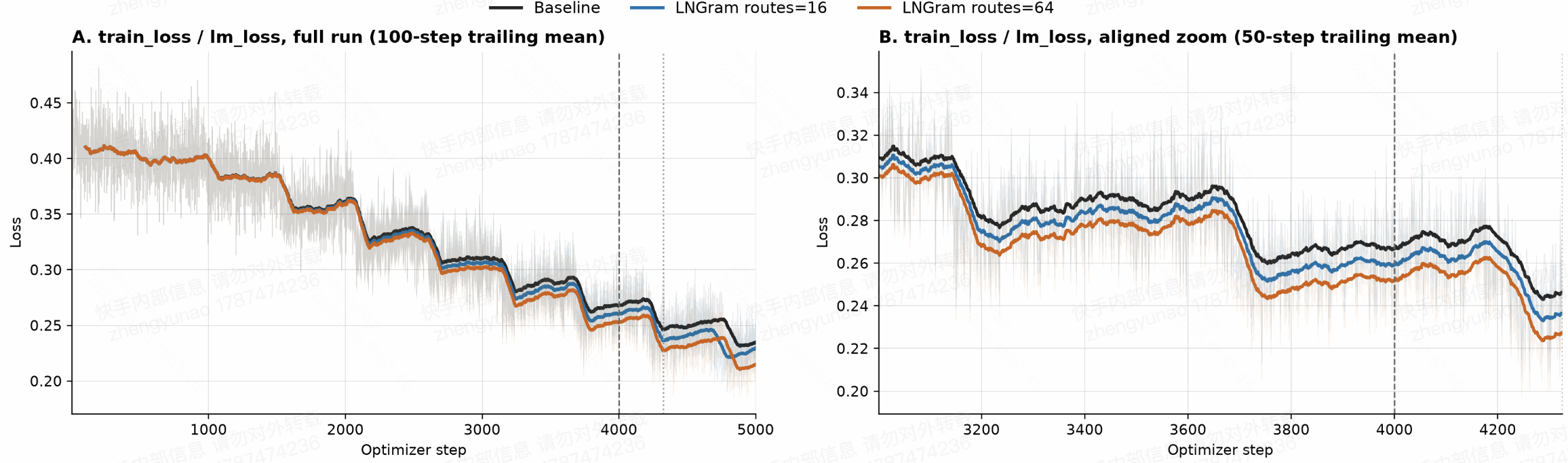}
    \caption{Comparison of loss.}
    \label{fig:vs_ste}
\end{figure}

As shown in Figure~\ref{fig:counterfactual_vs_ste}, the training loss with the Lngram v2 surrogate gradient decreases consistently throughout training, whereas STE quickly stalls at a substantially higher loss. The same trend is observed on the validation set, where the two methods obtain validation losses of 2.8585 and 7.3349, respectively. These results indicate that the optimization difficulty of Lngram does not arise solely from the binary threshold itself, but also from the discontinuous changes in memory contents associated with different discrete addresses. Direct STE ignores how switching addresses changes the actual lookup result and therefore fails to provide the router with an effective signal for discrete selection. In contrast, the Lngram v2 surrogate gradient explicitly incorporates the memory changes induced by different candidate addresses, aligning the gradient with the actual forward behavior. Under this setting, direct STE fails to converge effectively, demonstrating that the surrogate gradient is an important component for stable training of Lngram v2.

\section{Adding Lngram v2 to Models with Stabilized Representations}
\label{app:lngram_stability}

This section analyzes an empirical observation: Lngram v2 provides consistent gains when added to the mature Keye2B VLM, whereas its gains are weaker in a VLM assembled from pretrained SigLIP2 and Qwen3-1.7B-Instruct that is still undergoing cross-modal adaptation. As shown in Table~\ref{tab:vlm_lngram1}, adding Lngram v2 to Keye2B and training for only 5B tokens yields substantially larger gains than those obtained after 40B tokens of training for the model assembled from SigLIP2 and Qwen3-1.7B-Instruct. Further analysis suggests that this difference mainly arises because the continuous representations of the backbone continue to evolve during cross-modal adaptation, making the discrete addresses learned by Lngram v2 difficult to reuse consistently throughout training. This instability in the address space further limits the learning of memory representations, leaving the readout with only a weak influence on the backbone hidden states and final predictions.

\begin{table*}[t]
\centering
\caption{Comparison of Qwen3+SigLIP2 and Keye2B with Lngram v2.}
\label{tab:vlm_lngram1}
\resizebox{\textwidth}{!}{
\begin{tabular}{lccccccccccc}
\toprule
Model & OCR & MathVista & MMVet & Hallusion & MMStar & MMMU & AI2D & MMB-EN & MMB-CN & Bench V2.1 & Avg \\
\midrule
Keye2B
& 65.30 & 51.00 & 36.74 & 60.78 & 44.73 & \textbf{33.89} & 59.00 & 52.09 & 32.12 & 35.57 & 47.12 \\

Keye2B+Lngram v2
& \textbf{66.70} & \textbf{51.50} & \textbf{38.44} & \textbf{61.41} & \textbf{45.53} & 33.00 & \textbf{60.14} & \textbf{52.55} & \textbf{33.13} & \textbf{35.79} & \textbf{47.82} \\
\midrule

Qwen3+SigLIP2
& 48.20 & 41.01 & 26.65 & \textbf{38.20} & 62.50 & \textbf{41.53} & \textbf{42.56} & 64.63 & \textbf{63.62} & 43.17 & 47.21 \\

Qwen3+SigLIP2+Lngram v2
& \textbf{48.80} & \textbf{41.43} & \textbf{28.44} & 38.10 & \textbf{63.76} & \textbf{41.53} & 39.67 & \textbf{65.33} & 63.24 & \textbf{43.54} & \textbf{47.38} \\
\bottomrule
\end{tabular}
}
\end{table*}

\subsection{Experimental Setup}

We deterministically select 48 documents from a fixed subset of COCO and keep the images, text prompts, and input templates unchanged across checkpoints and intervention conditions. All inputs use the same question,
\texttt{Describe this image briefly.},
without providing captions, categories, or answer information.

Because the two model families use different visual tokenization schemes, the number of visual tokens differs across models. We therefore first aggregate within each document and then perform document-level statistics for all cross-model comparisons, rather than attempting token-wise alignment across models. All confidence intervals are obtained from 10,000 document-level bootstrap resamples using a fixed random seed.

We evaluate address stability in two ways. The first uses the router from each checkpoint itself. The second fixes the router from the final checkpoint of the same model family and disables all Lngram v2 outputs, thereby isolating the effect of changes in backbone hidden states on the discrete addresses. The latter separates changes in the router parameters from changes in the input representations and is therefore used as our primary analysis.

\subsection{Cross-Checkpoint Stability of Discrete Addresses}

Lngram v2 accesses its memory tables through exact discrete n-gram addresses. If the same semantics are repeatedly mapped to different addresses during training, the same table entry cannot consistently receive coherent training signals. We therefore compare route-code agreement, bit-level Hamming distance, and 2-gram and 3-gram address survival rates between adjacent checkpoints.

Table~\ref{tab:lngram_address_stability} reports the results when the final router is fixed and all Lngram v2 outputs are disabled. For visual tokens, the newly assembled VLM exhibits substantially lower address stability than the mature Keye2B model, and the gap becomes even larger at Layer 13. For example, visual route-code agreement at Layer 13 decreases from $0.9246$ for Keye2B to $0.7702$ for the new VLM, while 3-gram survival decreases from $0.7945$ to $0.4754$.

\begin{table*}[t]
\centering
\small
\caption{Discrete address stability of Lngram v2.}
\label{tab:lngram_address_stability}
\begin{tabular}{c c c c c c}
\toprule
Layer & Modality & Code Agreement $\uparrow$ & Hamming $\downarrow$ &
2-gram Survival $\uparrow$ & 3-gram Survival $\uparrow$ \\
\midrule
1  & Visual & 0.8859 / 0.9883 & 0.1204 / 0.0118 & 0.7900 / 0.9769 & 0.7079 / 0.9657 \\
1  & Text   & 0.9249 / 0.9462 & 0.0774 / 0.0575 & 0.8578 / 0.9028 & 0.7964 / 0.8625 \\
13 & Visual & 0.7702 / 0.9246 & 0.2630 / 0.0779 & 0.6020 / 0.8565 & 0.4754 / 0.7945 \\
13 & Text   & 0.8971 / 0.9163 & 0.1081 / 0.0866 & 0.8072 / 0.8356 & 0.7254 / 0.7616 \\
\bottomrule
\end{tabular}
\end{table*}

Document-level paired bootstrap yields the same conclusion. Relative to Keye2B, visual code agreement at Layers 1 and 13 decreases by $0.1024$ and $0.1544$, respectively, with corresponding 95\% CIs of
\[
[-0.1035,-0.1013]
\quad\text{and}\quad
[-0.1568,-0.1519].
\]
The corresponding decreases in 3-gram survival are $0.2579$ and $0.3190$. The text modality exhibits differences in the same direction, but with substantially smaller magnitudes.

The same trend is reproduced when each checkpoint uses its own router. For example, at Layer 13, visual code agreement / 3-gram survival is $0.7122/0.3813$ in the new model, compared with $0.8818/0.6964$ in the mature Keye2B model. This phenomenon therefore cannot be explained solely by changes in the router parameters and instead primarily reflects changes in the continuous representations fed into Lngram v2 over the course of training.

These results show that the stability of continuous representations directly affects memory modules that rely on exact discrete addresses. If the same local pattern frequently crosses discrete boundaries during training, semantically similar training examples are distributed across different table entries, making it difficult for any individual memory address to accumulate consistent information over time.

\subsection{Layer-Wise Propagation of Lngram v2 Residuals}

Address instability further increases the difficulty of learning effective Lngram v2 memory representations. To analyze their actual influence on the backbone representations, we separately disable Lngram v2 at Layer 1, Layer 13, or both layers, and compare the hidden states before and after the interventions.

We define the normalized representation difference at layer $l$ as
\[
R_l
=
\frac{
\lVert h_l^{\mathrm{on}}-h_l^{\mathrm{off}}\rVert_2
}{
\lVert h_l^{\mathrm{on}}\rVert_2
}.
\]

To further distinguish between ``small initial injection'' and ``attenuation during subsequent propagation,'' we define
\[
\mathrm{Retention}_l
=
\frac{
\lVert \Delta h_l\rVert_2
}{
\lVert \delta_{\mathrm{inj}}\rVert_2
},
\]
where $\delta_{\mathrm{inj}}$ denotes the effective Lngram v2 perturbation that actually enters the hidden state after the BF16 residual addition.

The results are shown in Table~\ref{tab:lngram_propagation}. The most prominent difference in the new model is not that Lngram v2 perturbations decay more rapidly in subsequent layers, but rather that their initial injections are substantially smaller. In particular, at Layer 13, the immediate $R$ at visual positions is $5.99\times10^{-3}$, approximately $1/9.6$ of that in the mature model; at text positions it is only $9.26\times10^{-5}$, approximately $1/68.7$ of the mature-model value.

\begin{table*}[t]
\centering
\small
\caption{Immediate injection magnitude and final propagation of Lngram v2 interventions.}
\label{tab:lngram_propagation}
\begin{tabular}{c c c c c c}
\toprule
Layer & Modality & Model &
Immediate $R$ & Final $R$ & Final Retention \\
\midrule
1  & Visual & New     & 0.01465   & 0.06442 & 16.64 \\
1  & Visual & Keye2B  & 0.01841   & 0.09515 & 3.685 \\
1  & Text   & New     & 0.008512  & 0.01281 & 12.79 \\
1  & Text   & Keye2B  & 0.04306   & 0.11646 & 4.935 \\
13 & Visual & New     & 0.005987  & 0.02476 & 3.109 \\
13 & Visual & Keye2B  & 0.05771   & 0.10009 & 1.890 \\
13 & Text   & New     & 0.0000926 & 0.01069 & 3.432 \\
13 & Text   & Keye2B  & 0.006358  & 0.11245 & 2.356 \\
\bottomrule
\end{tabular}
\end{table*}

To determine whether these representation differences ultimately affect model predictions, we additionally disable both Lngram v2 layers simultaneously and compare the final hidden states and next-token logits.

\begin{table}[t]
\centering
\small
\caption{Changes in final predictions after disabling all Lngram v2 layers.}
\label{tab:lngram_logit_effect}
\begin{tabular}{lcc}
\toprule
Metric & New VLM & Keye2B \\
\midrule
Final hidden $R$ (Visual) & 0.06510 & 0.12395 \\
Final hidden $R$ (Text)   & 0.01303 & 0.14383 \\
Next-token logit $R$      & 0.01741 & 0.19282 \\
Next-token JS             & 0.000297 & 0.01250 \\
Top-1 agreement           & 1.000 & 0.875 \\
\bottomrule
\end{tabular}
\end{table}

In the mature Keye2B model, disabling Lngram v2 results in a normalized next-token logit change of $0.1928$ and a Jensen--Shannon (JS) divergence of $1.25\times10^{-2}$, while changing the next-token top-1 prediction for $6/48$ documents. In contrast, the new model exhibits a logit $R$ of only $0.0174$ and a JS divergence of $2.97\times10^{-4}$, with the top-1 prediction unchanged for all 48 documents.

This shows that the two model families differ not only in address stability but also in functional allocation: the mature Keye2B model has already assigned a substantial predictive role to Lngram v2, whereas Lngram v2 in the new model still has only a weak influence on the final predictions.

We further examine the directional relationship between the Lngram v2 injection and the attention/MLP residual at the same layer. The cosine similarities are close to zero in both model families, with mean absolute values no greater than approximately $0.016$. Thus, there is no evidence that the Lngram v2 output merely duplicates the backbone residual.

At the same time, local counteracting components of the backbone residual along the direction of the Lngram v2 injection can indeed be observed in some layers of the new model. However, these components are insufficient to cancel the overall perturbation: the total hidden-state difference induced by the Lngram v2 intervention is preserved or amplified in subsequent layers. Local compensation may therefore exist, but it cannot explain the weaker final gains observed in the new model.

\subsection{Summary}

Taken together, the experiments above yield two relatively independent observations.

First, the visual discrete addresses of the newly assembled VLM are substantially less stable during training, particularly at the deeper Layer 13. Because Lngram v2 accesses its memory tables using exact discrete n-gram addresses, this non-stationarity reduces the probability that the same address continues to correspond to similar representations over time, dispersing training signals across different table entries and making stable memory formation more difficult.

This in turn leads to substantially smaller Lngram v2 injection magnitudes and predictive functional loads in the new model than in the mature Keye2B model. Existing Lngram v2 perturbations are not rapidly eliminated by the subsequent backbone layers; rather, the key difference is that optimization never assigns Lngram v2 a functional weight comparable to that learned in the mature model.

Thus, using pretrained vision and language models alone is not sufficient to guarantee that Lngram v2 can immediately become effective. Although both unimodal components already possess strong representational capabilities, the deep hidden space on which Lngram v2 relies may still change substantially while the cross-modal scaffold continues to adapt, which is unfavorable for forming stable discrete memory.

Based on this observation, in our formal comparison experiments we introduce Lngram v2 only after the model has entered the middle stage of training and its cross-modal representations have become more stable, rather than adding the module from the beginning of pretraining.

\section{Supplementary Analysis of Lngram Discrete IDs}
\label{app:id_analysis}

This section supplements the discrete representation analysis in Section~\ref{sec:analysis}, including statistics on discrete-code utilization, the statistical protocol and representative examples for route-code semantic associations, and detailed experiments on distributed semantic representation across multiple routes.

\subsection{Discrete-Code Utilization}
\label{app:code_health}

Before using discrete IDs for semantic analysis, we first verify that the Lngram discrete space does not exhibit substantial collapse. For each route, we compute the entropy of its code-value distribution as
\begin{equation}
H=-\sum_c p(c)\log p(c),
\qquad
N_{\mathrm{eff}}=\exp(H),
\end{equation}
where $N_{\mathrm{eff}}$ denotes the effective number of code values under the corresponding distribution.

Table~\ref{tab:code_health} reports statistics for the two analyzed layers. Each route uses approximately eight effective code values on average, with a normalized entropy of approximately 0.72 and an average frequency of approximately 30\% for the most common code. Moreover, every theoretically possible route code is used in both layers, with no dead codes.

\begin{table}[t]
\centering
\caption{Utilization statistics of Lngram v2 discrete codes.}
\label{tab:code_health}
\begin{tabular}{lcccc}
\toprule
Layer & Effective Codes & Normalized Entropy & Top-Code Frequency & Dead Codes \\
\midrule
1  & 7.7916 & 0.7210 & 0.3206 & 0 / 2,048 \\
13 & 7.9979 & 0.7291 & 0.3093 & 0 / 2,048 \\
\bottomrule
\end{tabular}
\end{table}

Therefore, the ID-only semantic readout capability observed in the main text does not arise from degeneration of the discrete space into a small number of frequently used codes, but instead emerges while the discrete address space remains broadly utilized.

\subsection{Route-Code Semantic Associations and Representative Examples}
\label{app:id_dictionary}

Beyond the complete signature, we further examine whether semantic information can be localized to individual route codes. This analysis is related in spirit to prior work that identifies concept-level semantics in internal neural units~\citep{bau2017network}. However, the objects analyzed here are not separately trained probes or dictionary units, but the discrete addresses natively generated by the Lngram memory mechanism.

Let $C_{\ell,r}$ denote the discrete code of route $r$ at layer $\ell$. We denote a code--semantic relation as
\begin{equation}
C_{\ell,r}=c \Rightarrow z,
\end{equation}
where $z$ is the semantic concept under analysis. On the held-out set, we define
\begin{equation}
P
=
P_{\mathrm{heldout}}
\left(
z\mid C_{\ell,r}=c
\right),
\end{equation}
and let $P_0$ denote the baseline probability after controlling for nuisance factors such as token position and image source. The corresponding increase in semantic probability is
\begin{equation}
\Delta=P-P_0.
\end{equation}

All formal statistical associations follow a strict reference--held-out separation protocol. Candidate code--semantic relations and their effect directions are determined exclusively on the reference data and are subsequently tested independently on the held-out data. The reference set yields 16,304 candidate associations, of which 14,002 can be evaluated on the held-out set. Ultimately, 9,220 satisfy the requirements for consistent effect direction, confidence intervals, and Benjamini--Hochberg multiple-testing correction~\citep{benjamini1995controlling}, corresponding to an overall replication rate of 65.85\%. For local presence, the replication rates at Layers 1 and 13 reach 83.01\% and 84.94\%, respectively.

To provide a more intuitive view of the semantic structure encoded by the discrete codes, we further list 15 representative examples selected from these analyzable associations. The displayed entries have relatively high held-out conditional probabilities, with positive probability increases on both the reference and held-out sets. Importantly, this selection is used only for qualitative illustration of intuitive code--semantic correspondences and is not used to compute the replication rates or statistical-significance results above. The formal statistical evidence continues to come from reference--held-out validation over the complete candidate set.

\begin{table*}[t]
\centering
\caption{Representative semantic correspondences of Lngram route codes. $P$ denotes the conditional probability of the semantic concept given the code on the held-out set, $P_0$ denotes the baseline probability after controlling for nuisance factors such as position and image source, and $\Delta=P-P_0$.}
\label{tab:dictionary_examples}
\setlength{\tabcolsep}{8pt}
\renewcommand{\arraystretch}{1.06}
\begin{tabular}{llccc}
\toprule
Code Expression & Semantic Concept & $P$ & $P_0$ & $\Delta$ \\
\midrule
$C_{1,78}=8$    & skis          & 11.03\% & 2.90\%  & +8.13 pp \\
$C_{13,53}=14$  & person        & 63.25\% & 53.60\% & +9.65 pp \\
$C_{1,47}=11$   & cat           & 9.84\%  & 3.97\%  & +5.87 pp \\
$C_{1,6}=15$    & dining table  & 14.65\% & 9.67\%  & +4.98 pp \\
$C_{13,97}=15$  & kite          & 11.60\% & 2.07\%  & +9.53 pp \\
$C_{1,63}=0$    & car           & 21.03\% & 11.13\% & +9.89 pp \\
$C_{1,117}=4$   & skateboard    & 9.57\%  & 2.80\%  & +6.77 pp \\
$C_{1,75}=10$   & giraffe       & 21.70\% & 2.37\%  & +19.33 pp \\
$C_{13,83}=9$   & chair         & 13.38\% & 10.07\% & +3.31 pp \\
$C_{1,48}=5$    & sports ball   & 9.36\%  & 3.33\%  & +6.03 pp \\
$C_{1,71}=9$    & traffic light & 11.75\% & 3.30\%  & +8.45 pp \\
$C_{1,75}=9$    & horse         & 10.06\% & 2.23\%  & +7.83 pp \\
$C_{1,73}=0$    & airplane      & 9.71\%  & 3.03\%  & +6.68 pp \\
$C_{1,75}=11$   & zebra         & 10.62\% & 1.63\%  & +8.99 pp \\
$C_{13,106}=7$  & surfboard     & 9.39\%  & 2.47\%  & +6.92 pp \\
\bottomrule
\end{tabular}
\end{table*}

Table~\ref{tab:dictionary_examples} shows several discrete codes with intuitive semantic interpretations. For example, when
\[
C_{1,75}=10 \Rightarrow \text{giraffe},
\]
the conditional probability of giraffe on the held-out set reaches 21.70\%, substantially above the corresponding baseline of 2.37\%, representing an increase of 19.33 percentage points. Similarly,
\[
C_{13,53}=14 \Rightarrow \text{person}
\]
corresponds to a held-out conditional probability of 63.25\%, which is 9.65 percentage points above the baseline.

These correspondences span diverse concepts, including people, animals, vehicles, and common objects. Notably, different codes within the same route can also exhibit distinct but related semantics. For example, codes 9, 10, and 11 on route 75 show strong positive associations with horse, giraffe, and zebra, respectively. This indicates that a route itself does not simply correspond to one fixed concept; rather, its discrete values can distinguish different semantic states within the same local discrete space.

Thus, Lngram's discrete IDs not only preserve the semantic structure of hidden states at the level of complete signatures, but individual route codes can also form statistically reproducible associations with clearly interpretable concepts. The originally continuous and difficult-to-enumerate internal hidden states can therefore be transformed into queryable discrete semantic relations of the form
\[
C_{\ell,r}=c \Rightarrow z.
\]

\subsection{Detailed Analysis of Distributed Semantics}
\label{app:distributed_semantics}

Individual route codes already exhibit stable semantic tendencies, yet the complete IDs provide substantially stronger readout capability in Table~\ref{tab:id_semantic} of the main text. To directly determine whether this difference arises from joint encoding across multiple routes, we recombine the single-code semantic effects estimated on the reference set.

For a complete signature
\[
C=(c_1,\ldots,c_{128})
\]
and a concept $z$, we define
\begin{equation}
S_{\mathrm{dict}}(C,z)
=
\frac{1}{128}
\sum_{r=1}^{128}
\Delta_{\mathrm{ref}}(r,c_r,z),
\end{equation}
where $\Delta_{\mathrm{ref}}(r,c_r,z)$ denotes the semantic effect of the current code $c_r$ on route $r$ for concept $z$, as estimated from the reference data. For single-code mappings that do not satisfy the statistical-support criteria on the reference set, the corresponding effect is set to zero.

On the held-out set, this reader uses only discrete IDs, without accessing any continuous hidden states or refitting parameters on the held-out data. Its performance therefore directly measures whether route-code semantic associations independently identified on the reference set can be composed into semantic representations that generalize to unseen samples.

Simply summing the independent single-code effects significantly outperforms the nuisance baseline in all six experimental settings and retains 31.79\%--54.82\% of the excess AP of $H_{\mathrm{model}}$. This demonstrates that the weak semantic signals present in individual route codes are not isolated phenomena, but can be combined into effective semantic predictions.

To further analyze the complementary contributions of different routes, for each query--concept pair we rank routes in descending order according to
\begin{equation}
|\Delta_{\mathrm{ref}}(r,c_r,z)|
\end{equation}
and sequentially construct semantic predictions using
\[
m\in\{1,2,4,8,16,32,64,128\}
\]
routes.

Across all six experimental settings, we observe
\begin{equation}
AP_{128}>AP_1,
\end{equation}
with
\begin{equation}
\frac{AP_{128}}{AP_1}
=
2.02\text{--}12.32.
\end{equation}

Here, $AP_1$ already selects the single route with the largest contribution for each query--concept pair. Therefore, the advantage of the complete signature cannot be explained by one consistently dominant route and must instead arise from complementary information provided by multiple routes.

This result also explains the performance gap between the single-code semantic dictionary and the complete-ID reader. The reader based on independent route-code marginal effects retains only 31.79\%--54.82\% of the excess AP of the continuous representation, whereas ID-KNN operating directly on the complete signature retains 65.77\%--84.27\%. Thus, the marginal semantics of individual route codes explain only part of the information in the complete discrete representation; the joint configuration of multiple routes additionally encodes compositional semantic structure that cannot be fully recovered from independent single-code effects.

In summary, Lngram's discrete representations provide two complementary levels of interpretability: individual route codes establish stable and reproducible semantic correspondences with specific concepts, while the complete signature preserves richer hidden-state semantics through distributed combinations across multiple routes. This enables Lngram IDs to serve both as variables for exact memory addressing and as a structured discrete interface for analyzing the model's internal continuous representations.


\section{Supplementary Experimental Results}
\label{app:experimental_details}

\subsection{Complete Statistical Tests for Keye30B}
\label{app:vlm_significance}

We conduct paired statistical tests to assess the reliability of the Keye30B results. We compare the Baseline and Lngram v2 (routes=256) checkpoints at training step 5,000, each evaluated once on the same ten benchmark datasets and splits using identical benchmark definitions and deterministic decoding settings. The reported scores are obtained from the frozen evaluation outputs. For statistical testing, predictions are aligned by sample identifier, and scorer-produced per-sample scores before rounding are used. We further verify that aligned samples have identical reference answers and benchmark metadata. Thus, all differences and uncertainty estimates are computed at the paired sample level rather than from rounded aggregate scores.

For benchmarks with binary per-sample outcomes, we use the two-sided exact paired McNemar test. MMVet provides fractional per-sample scores and is analyzed using a paired sign-flip permutation test. For HallusionBench, where multiple questions may share the same image, permutation testing and bootstrap resampling are performed over 346 image clusters rather than 951 individual questions. For MMMU, we use the 900 examples in the official validation split and construct confidence intervals using category-stratified paired bootstrap. For MMBench, $n=1{,}292$ corresponds to the unique scored questions after aggregation of circularized inference rows. All remaining confidence intervals use ordinary paired bootstrap. We use 200,000 bootstrap replicates for 95\% confidence intervals and 1,000,000 sign-flip draws for MMVet and HallusionBench. The ten per-benchmark two-sided $p$-values are corrected using both Holm~\citep{holm1979simple}, which controls the family-wise error rate, and Benjamini--Hochberg (BH)~\citep{benjamini1995controlling}, which controls the false discovery rate.

\begin{table*}[t]
\centering
\caption{Paired statistical tests for the step-5,000 Keye30B Baseline and Lngram v2 checkpoints. Scores, differences, and confidence intervals are reported in percentage points.}
\label{tab:vlm_significance}
\scriptsize
\setlength{\tabcolsep}{3.5pt}
\resizebox{\textwidth}{!}{
\begin{tabular}{lrrrrrrrr}
\toprule
Benchmark & $n$ & Baseline & Lngram v2 & $\Delta$ & 95\% CI & Raw $p$ & Holm $p$ & BH $q$ \\
\midrule
OCRBench
& 1,000 & 81.30 & 81.00 & -0.30 & $[-1.90,+1.30]$ & 0.8043 & 1.0000 & 0.8937 \\
MathVista
& 1,000 & 80.90 & 80.80 & -0.10 & $[-2.00,+1.80]$ & 1.0000 & 1.0000 & 1.0000 \\
MMVet
& 218 & 69.45 & 72.02 & +2.57 & $[-0.69,+5.96]$ & 0.1398 & 0.6991 & 0.2330 \\
HallusionBench
& 951 & 74.03 & 74.87 & +0.84 & $[-1.46,+3.16]$ & 0.5312 & 1.0000 & 0.6640 \\
MMStar
& 1,500 & 69.60 & 70.93 & +1.33 & $[-0.60,+3.27]$ & 0.1939 & 0.7756 & 0.2770 \\
MMMU
& 900 & 67.78 & 70.22 & +2.44 & $[0.00,+4.89]$ & 0.0672 & 0.4702 & 0.1679 \\
AI2D
& 3,088 & 82.09 & 83.97 & +1.88 & $[+0.68,+3.08]$ & 0.0023 & 0.0210 & 0.0117 \\
MMBench-EN
& 1,292 & 83.75 & 86.92 & +3.17 & $[+1.86,+4.49]$
& $4.19\times10^{-6}$ & $4.19\times10^{-5}$ & $4.19\times10^{-5}$ \\
MMBench-CN
& 1,292 & 81.73 & 83.75 & +2.01 & $[+0.54,+3.48]$ & 0.0088 & 0.0702 & 0.0293 \\
Benchmark V2.1
& 1,355 & 91.66 & 92.62 & +0.96 & $[-0.07,+1.99]$ & 0.0919 & 0.5515 & 0.1838 \\
\midrule
10-task Average
& -- & 78.23 & 79.71 & +1.48 & $[+0.86,+2.11]$
& $5.00\times10^{-6}$ & -- & -- \\
\bottomrule
\end{tabular}
}

\vspace{2pt}
\begin{minipage}{0.98\textwidth}
\footnotesize
\end{minipage}
\end{table*}

Lngram v2 improves the equally weighted ten-task average by 1.48 percentage points, with a fixed-suite 95\% paired-bootstrap confidence interval of $[0.86,2.11]$ and a two-sided paired-randomization $p$-value of $5.00\times10^{-6}$. The improvement is also consistent across benchmarks: Lngram v2 outperforms the Baseline on eight of the ten tasks. An exact sign-flip test over the ten task effects, which accounts for both their directions and magnitudes, yields $p=0.0078$. Together, these results show that the overall gain is not driven by a small number of isolated benchmarks, but reflects a broadly positive effect across the fixed evaluation suite.

At the individual-benchmark level, AI2D and MMBench-EN remain statistically significant after the conservative Holm correction, while MMBench-CN is additionally significant under BH correction. In particular, Lngram v2 achieves gains of +1.88, +3.17, and +2.01 percentage points on AI2D, MMBench-EN, and MMBench-CN, respectively. These results provide strong evidence that Lngram v2 yields reliable improvements on Keye30B, with both a significant aggregate gain and clear improvements on several challenging multimodal benchmarks. 

\subsection{Complete Inference-Efficiency Results}
\label{app:efficiency_detail}

We evaluate the inference efficiency of different Lngram v2 configurations on a single H800 GPU. Prefill is measured using input lengths of 256, 512, and 1024, while the decode stage reports single-token latency. Table~\ref{tab:lngram_v2_efficiency} presents the complete results.

\begin{table*}[t]
\centering
\caption{Complete inference-efficiency results on a single H800 GPU. The three prefill values correspond to sequence lengths of 256, 512, and 1024, respectively.}
\label{tab:lngram_v2_efficiency}
\scriptsize
\setlength{\tabcolsep}{5pt}
\resizebox{\textwidth}{!}{
\begin{tabular}{lcccc}
\toprule
Configuration
& Prefill Latency (ms) $\downarrow$
& Prefill Memory (GiB) $\downarrow$
& Decode Latency (ms/token) $\downarrow$
& Decode Memory (GiB) $\downarrow$ \\
\midrule
Baseline
& 54.62 / 54.98 / 56.99
& 4.90 / 5.01 / 5.24
& 48.26 & 4.89 \\
r16/m4/KV8
& 58.98 / 58.92 / 59.08
& 4.96 / 5.07 / 5.31
& 50.97 & 4.95 \\
r64/m4/KV8
& 57.84 / 57.86 / 60.49
& 5.08 / 5.21 / 5.46
& 52.14 & 5.05 \\
r23/m4/KV8/mem256
& 58.03 / 58.59 / 60.85
& 5.06 / 5.17 / 5.40
& 51.79 & 5.05 \\
r512/m4/KV8/mem256
& 57.88 / 57.59 / 83.75
& 8.67 / 10.34 / 13.68
& 51.44 & 7.09 \\
r16/m5/KV2
& 56.75 / 57.15 / 59.46
& 5.19 / 5.30 / 5.53
& 49.92 & 5.18 \\
r123/m4/KV2
& 57.21 / 57.59 / 59.96
& 5.19 / 5.30 / 5.53
& 50.55 & 5.18 \\
r121/m4/KV16
& 60.75 / 64.17 / 73.80
& 5.48 / 5.87 / 6.67
& 50.74 & 5.18 \\
\bottomrule
\end{tabular}
}
\end{table*}

The additional online computational overhead of standard configurations is generally modest. For example, for r64/m4/KV8 at sequence length 1024, prefill latency increases from 56.99 ms for the Baseline to 60.49 ms, corresponding to an overhead of approximately 6.1\%, while memory usage increases from 5.24 GiB to 5.46 GiB. Decode latency increases from 48.26 ms/token to 52.14 ms/token, corresponding to approximately 8.0\% overhead, while memory usage increases only from 4.89 GiB to 5.05 GiB.

In contrast, the extremely large r512/m4/KV8/mem256 configuration incurs substantially higher long-sequence memory usage and prefill overhead, consistent with the diminishing returns from capacity scaling observed in the ablation experiments in the main text. In practical deployments, moderate-scale configurations therefore provide a more favorable trade-off among downstream performance, parameter capacity, and online computation cost.

\subsection{Experimental Setup for the Controlled Comparison of Lngram v1 and v2}
\label{app:v1v2_setup}

The comparison between Lngram v1 and v2 uses the same 0.6B DeepSeek-V2-style sparse MoE model~\citep{deepseekai2024deepseekv2,dai2024deepseekmoe}. The model contains 24 Transformer layers with a hidden size of 768, 12 attention heads, and 2 KV heads. Each MoE layer contains 8 routed experts and 1 shared expert, with 2 routed experts activated per token.

Lngram is inserted into the 2nd and 12th layers in all variants and uses 4-bit discrete codes together with 2/3-gram memory. All models are trained from random initialization for 10,000 steps on the same FineWeb-Edu data~\citep{penedo2024fineweb}, processing approximately 1.31B tokens in total, with exactly the same data order, optimizer, and random seed. Therefore, the differences reported in Table~\ref{tab:v1_v2_comparison} of the main text primarily reflect the architectural differences between Lngram v1 and v2 rather than changes in training data or optimization settings.

\section{Hyperparameter Settings}
\label{sec:hyperparameters}

For reproducibility, Table~\ref{tab:lngram_hyperparameters} summarizes the main hyperparameters of Lngram v2 in the Keye2B and Keye30B comparison experiments. At both model scales, Lngram v2 is inserted into a shallow and an intermediate layer of the backbone, uses 4-bit discrete codes and 2/3-gram memory, and employs a multi-head readout to retrieve the memory representations.

\begin{table*}[t]
\centering
\caption{Lngram v2 hyperparameter settings for Keye2B and Keye30B.}
\label{tab:lngram_hyperparameters}
\resizebox{0.8\linewidth}{!}{
\begin{tabular}{lcc}
\toprule
Hyperparameter & Keye2B & Keye30B \\
\midrule
Insertion Layers & 2, 14 & 2, 24 \\
Number of Routes $R$ & 121 & 256 \\
Bits per Route & \multicolumn{2}{c}{4 bit} \\
Codes per Route & \multicolumn{2}{c}{16} \\
n-gram Orders & \multicolumn{2}{c}{2, 3} \\
Memory Dimension & 128 & 256 \\
Readout Q Heads & \multicolumn{2}{c}{16} \\
Readout KV Heads & \multicolumn{2}{c}{16} \\
Head Dimension & \multicolumn{2}{c}{128} \\
Readout Mode & full & streaming \\
Route Chunk Size & -- & 64 \\
Streaming Activation Checkpoint & -- & enabled \\
Sink & \multicolumn{2}{c}{initial effective-route mass $0.5$} \\
Router Surrogate Gradient & \multicolumn{2}{c}{approximate} \\
Lngram Dropout & \multicolumn{2}{c}{0} \\
Readout Attention Dropout & \multicolumn{2}{c}{0} \\
Short Convolution & \multicolumn{2}{c}{kernel$=4$, dilation$=3$, no bias, zero-init} \\
Table Initialization & \multicolumn{2}{c}{normal, std scale$=1.0$} \\
Output Projection Initialization & \multicolumn{2}{c}{zero-init} \\
Lngram Non-Table LR & \multicolumn{2}{c}{peak $1\times10^{-4}$, min $2\times10^{-6}$} \\
Table LR & \multicolumn{2}{c}{peak $4\times10^{-4}$, min $1\times10^{-5}$} \\
Table / Router-Q Weight Decay & \multicolumn{2}{c}{0} \\
Lngram Total Parameters & 155M & 595M \\
Steady-State Active Parameters / Token & 21M & 25M \\
\bottomrule
\end{tabular}
}
\end{table*}

The Keye30B experiment is trained on 8 nodes with a total of 64 GPUs. Tensor parallelism, pipeline parallelism, context parallelism, expert parallelism, and expert tensor parallelism are set to
\[
\mathrm{TP}=1,\quad
\mathrm{PP}=4,\quad
\mathrm{CP}=16,\quad
\mathrm{EP}=8,\quad
\mathrm{ETP}=1,
\]
respectively.
The sequence length is 32,768, with a micro batch size of 1 and a global batch size of 32, resulting in approximately $1.049$M tokens per training step.
The formal comparison uses the checkpoint at step 5000, corresponding to approximately $5.24$B cumulative training tokens.
Training uses BF16 with a random seed of 19260817. The base learning rate of the backbone is $5\times10^{-6}$ with a weight decay of 0.1, followed by cosine learning-rate decay after a linear warmup over the first 1,000 steps.

\section{Generative AI Use Statement}

Generative AI tools were used during the preparation of this work for language polishing and to assist with code development for experiments. All AI-assisted code was reviewed, tested, and validated by the authors before use. Generative AI was not used to generate, fabricate, or modify experimental results. All results reported in this paper were obtained from actual model training and evaluation runs and were verified by the authors. The authors take full responsibility for the content, methodology, implementation, and conclusions of this work.

\end{document}